\documentclass{article}

\PassOptionsToPackage{numbers, compress}{natbib}

\usepackage[final]{neurips_2019}

\usepackage[utf8]{inputenc} 
\usepackage[T1]{fontenc}    
\usepackage{hyperref}       
\usepackage{url}            
\usepackage{booktabs}       
\usepackage{amsfonts}       
\usepackage{nicefrac}       
\usepackage{microtype}      
\usepackage{url}
\usepackage{graphicx}
\usepackage{wrapfig}
\usepackage{subcaption}
\usepackage{listings}
\usepackage{array, booktabs, makecell}
\usepackage{xcolor}
\setcitestyle{square}
\usepackage[bottom]{footmisc}
\title{cFineGAN: Unsupervised multi-conditional fine-grained image generation}

\author{%
  Gunjan Aggarwal\thanks{Authors contributed equally} \\
  Adobe Inc, Noida, India\\
  \texttt{guaggarw@adobe.com} \\
   \And
   Abhishek Sinha$^{*}$\thanks{work done while author was working at Adobe, India} \\
   Stanford University \\
   \texttt{a7b23@stanford.edu}
   }

\begin{document}

\maketitle

\begin{abstract}
We propose an unsupervised multi-conditional image generation pipeline: cFineGAN, that can generate an image conditioned on two input images such that the generated image preserves the texture of one and the shape of the other input. To achieve this goal, we extend upon the recently proposed work of FineGAN \citep{singh2018finegan} and make use of standard as well as shape-biased pre-trained ImageNet models. We demonstrate both qualitatively as well as quantitatively the benefit of using the shape-biased network. We present our image generation result across three benchmark datasets- CUB-200-2011\citep{welinder2010caltech}, Stanford Dogs\citep{khosla2011novel} and UT Zappos50k\citep{finegrained}.

\end{abstract}

\section{Introduction}
\vspace{1pt}
Recent developments in deep learning and generative adversarial networks(GAN) have made it possible to generate realistic looking images of high resolution. The image generation techniques generally come in two forms :
i) Unconditional image generation – starting from a noise vector, the generator generates an
image \citep{goodfellow2014generative}.
ii) Conditional image generation – given a condition, the aim is to generate an image adhering to some condition \citep{mirza2014conditional, isola2017image, zhu2017unpaired}.

While a lot of work has been done in the domain of single-conditional image synthesis, the
domain of unsupervised multi-conditional image synthesis is relatively new. We aim to generate an image conditioned on two inputs such that the generated image contains texture of the first and shape of the second conditioned image.
\vspace{1pt}
\section{Approach}
\vspace{1pt}
Our work is based upon the recently released work FineGAN. The authors propose a GAN based framework that learns to disentangle the background, shape and texture of an
image in an unsupervised manner. The network generates an image conditioned on input background, shape and texture codes. 

Our pipeline takes in two images $I_1$ and $I_2$ as input and generates an output image(O). The pipeline consists of three steps - i) Compute the texture code(T) that describes the first input image($I_1$), ii) Compute the shape code(S) that describes the second input image($I_2$), and iii) Feed the computed codes(T and S) as input to the pre-trained FineGAN network to get the desired output O.

To compute the codes(T and S), we take a trained FineGAN and iterate over all the possible combinations of shape and texture codes for 10 different noise vectors(different noise vector lead to different orientations of the generated image). We denote as G the set of all such generated images.  
To compute the texture code(T), we compute the nearest neighbour of $I_1$ amongst G in the embedding space of ImageNet \citep{russakovsky2015imagenet} pre-trained ResNet50 model \citep{He_2016_CVPR}. The embedding space is defined by the Global Average Pooling layer output of the ResNet50 model.

We repeat the same process for image $I_2$ to compute the shape code(S) with the exception of using a shape biased pre-trained ResNet50 network. The motivation for using a shape biased network stems from \citep{geirhos2018imagenet}, where the authors show that the ImageNet trained models are biased towards texture details of image. The authors use stylized variants of the ImageNet dataset to train the network resulting in a shape-biased network. We hypothesize that the shape biasness of the network would allow it to better capture the shape details of the image, leading to correct identification of the shape code of an input image. We verify both quantitatively and qualitatively this design choice in the following section.  

\section{Results and Discussions}
\vspace{1pt}
\begin{figure}[htp]
    \centering
    \includegraphics[width=13.5cm, height = 3.5cm]{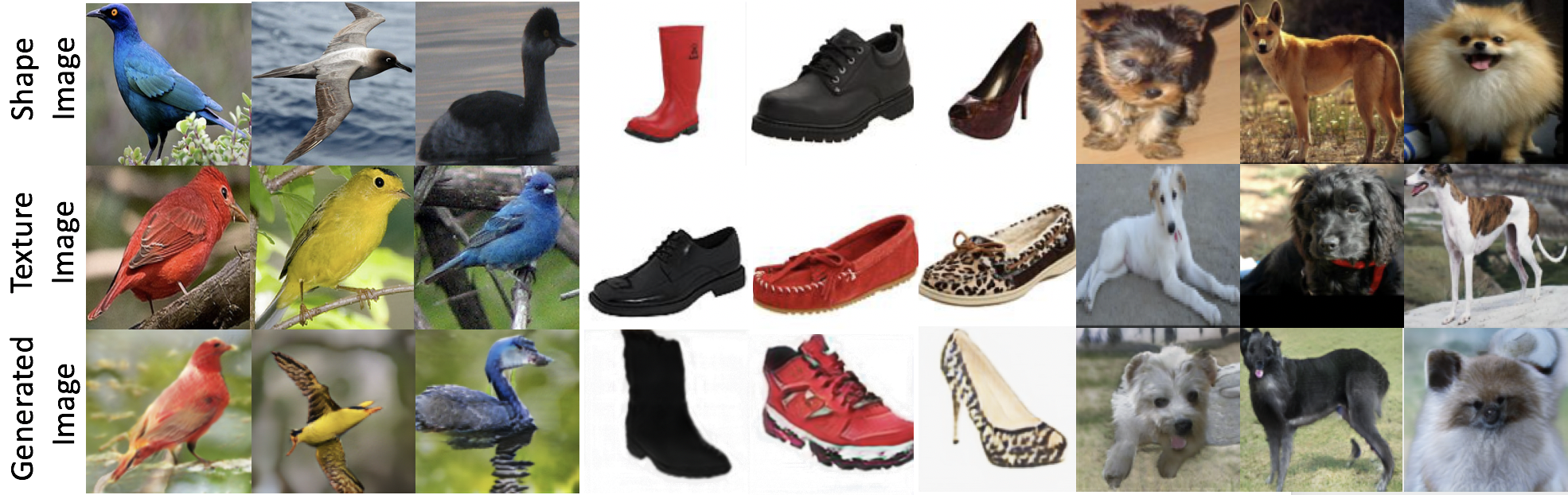}
    \caption{cFineGAN results - columns 1-3 show results for CUB-200-2011, 4-6 for UT Zappos50k and 7-9 for Stanford Dogs datasets respectively.}
    \vspace{1pt}
    \label{fig:results}
\end{figure}
Our cFineGAN results over the three datasets - CUB-200-2011 \citep{welinder2010caltech}, UT Zappos50k \footnote{We trained our own FineGAN model over this dataset} \citep{finegrained} and Stanford Dogs \citep{khosla2011novel} are shown in figure \ref{fig:results}. Additional results can be found in the Appendix section.

    


\begin{minipage}[b]{0.25\linewidth}
{\renewcommand{\arraystretch}{2}%
\begin{tabular}{ |c|c| } 
\hline
\thead{ResNet\\ Model} & \thead{Accuracy\\ (\%)}\\
 \hline
 \thead{Standard\\ResNet50\\} & 70.75 \\ 
 \thead{Shape-biased\\ResNet50\\} & 86.90 \\ 
 \hline
\end{tabular}} \quad
\medskip
\captionof{table}{Quantitative analysis of models for shape code prediction of generated images.}
\label{tab:shape-acc}
\end{minipage}
\hspace{0.03\linewidth}
\begin{minipage}[b]{0.33\linewidth}
\centering
\includegraphics[width=\textwidth, height =4.6cm]{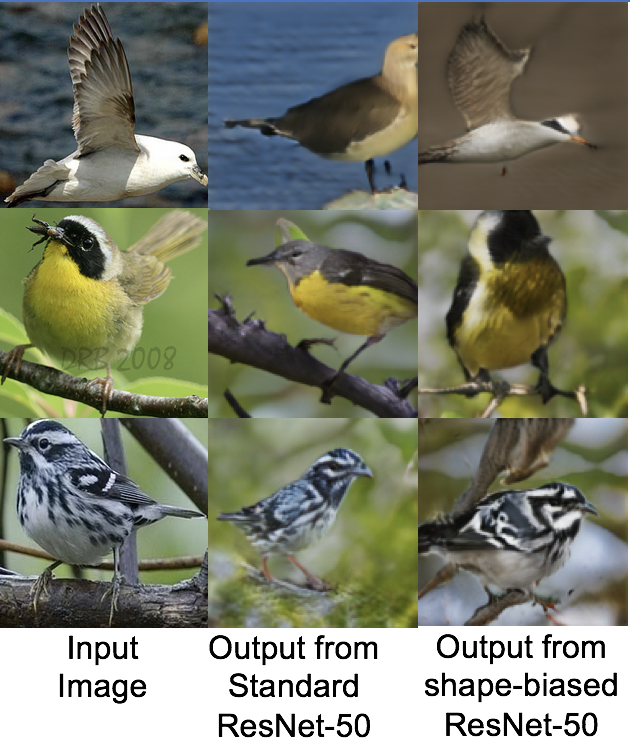}
\captionof{figure}{Nearest neighbour image for standard and shape-biased networks.}
\label{fig:figure1}
\end{minipage}
\hspace{0.01\linewidth}
\begin{minipage}[b]{0.37\linewidth}
\centering
\includegraphics[width=\textwidth, height =5cm]{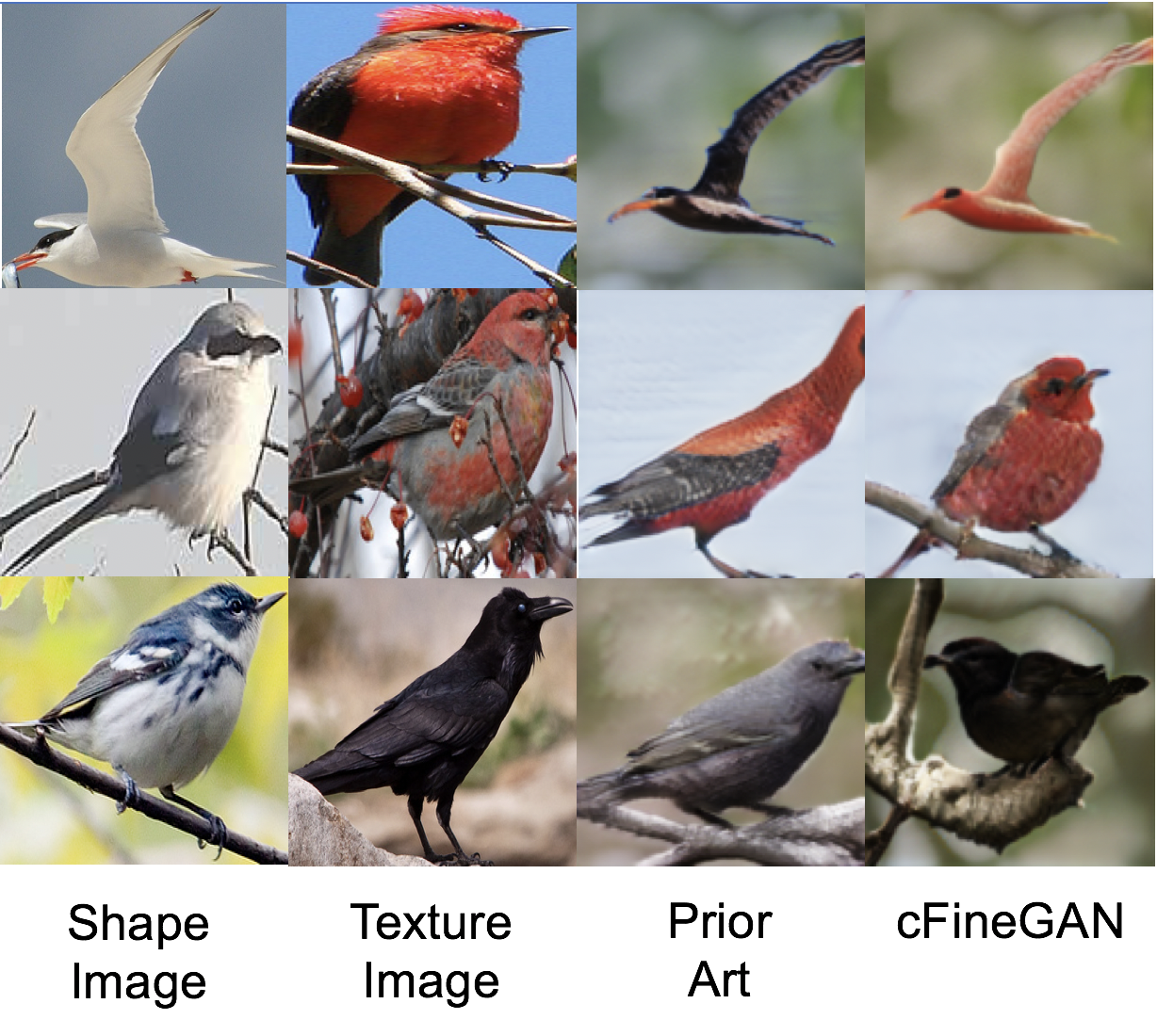}
\captionof{figure}{Qualitative comparison of cFineGAN against the prior art.}
\label{fig:figure2}
\end{minipage}

To quantitatively evaluate the benefit of using a shape-biased pre-trained model for extracting the shape code, we compute the nearest neighbour in the embedding space for each generated image in G. We define accuracy as the fraction of times the query image and its nearest neighbour have the same shape code. As the shape code of all the generated images is known, we can compute this metric. 
Table \ref{tab:shape-acc} shows that the accuracy achieved by the shape-biased model is much better than that of a standard model. Some qualitative results have been shown in figure \ref{fig:figure1}.

We baseline our method against the approach mentioned in \citep{singh2018finegan} where the authors train classifiers over the domain of generated images to predict the shape and texture codes given image as an input. Since the classifier is trained over the domain of generated images but is expected to predict the codes of natural images during evaluation time, the huge domain shift encountered between train and test settings lead to incorrect outputs. We show some qualitative comparisons against this baseline in figure \ref{fig:figure2}. As can be seen cFineGAN better captures the shape and texture details of input images.

  

{\small
\bibliographystyle{ieee}
\bibliography{egbib}
}

\newpage
\section{Appendix}
\label{appendix}

\subsection{Additional Results}

\begin{figure}[htp]
    \centering
    \includegraphics[width=13.5cm]{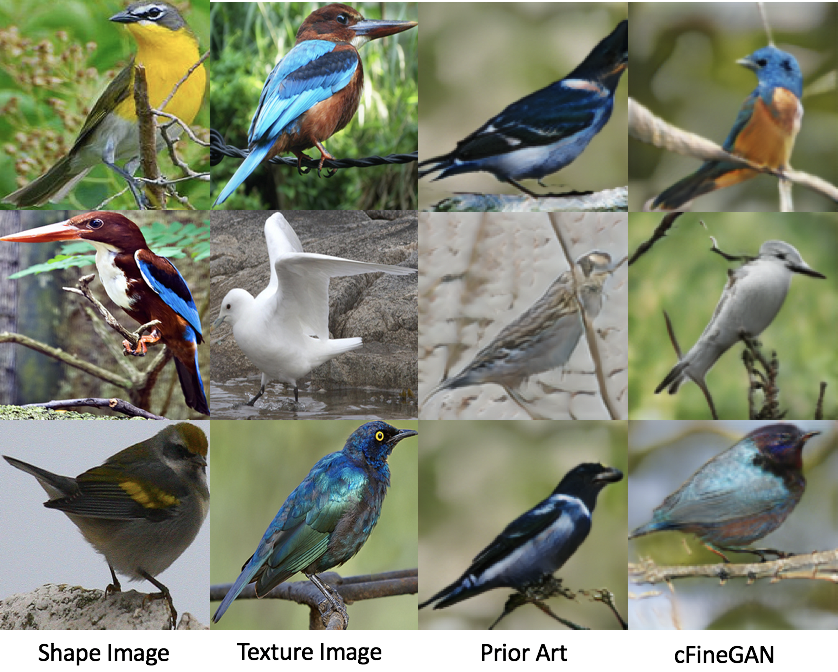}
    \captionsetup{font=large,labelfont=large}
    \caption{Additional results over the CUB-200-2011 dataset.}
    \label{fig:results_add_cub}
\end{figure}

\begin{figure}[htp]
    \centering
    \includegraphics[width=13.5cm]{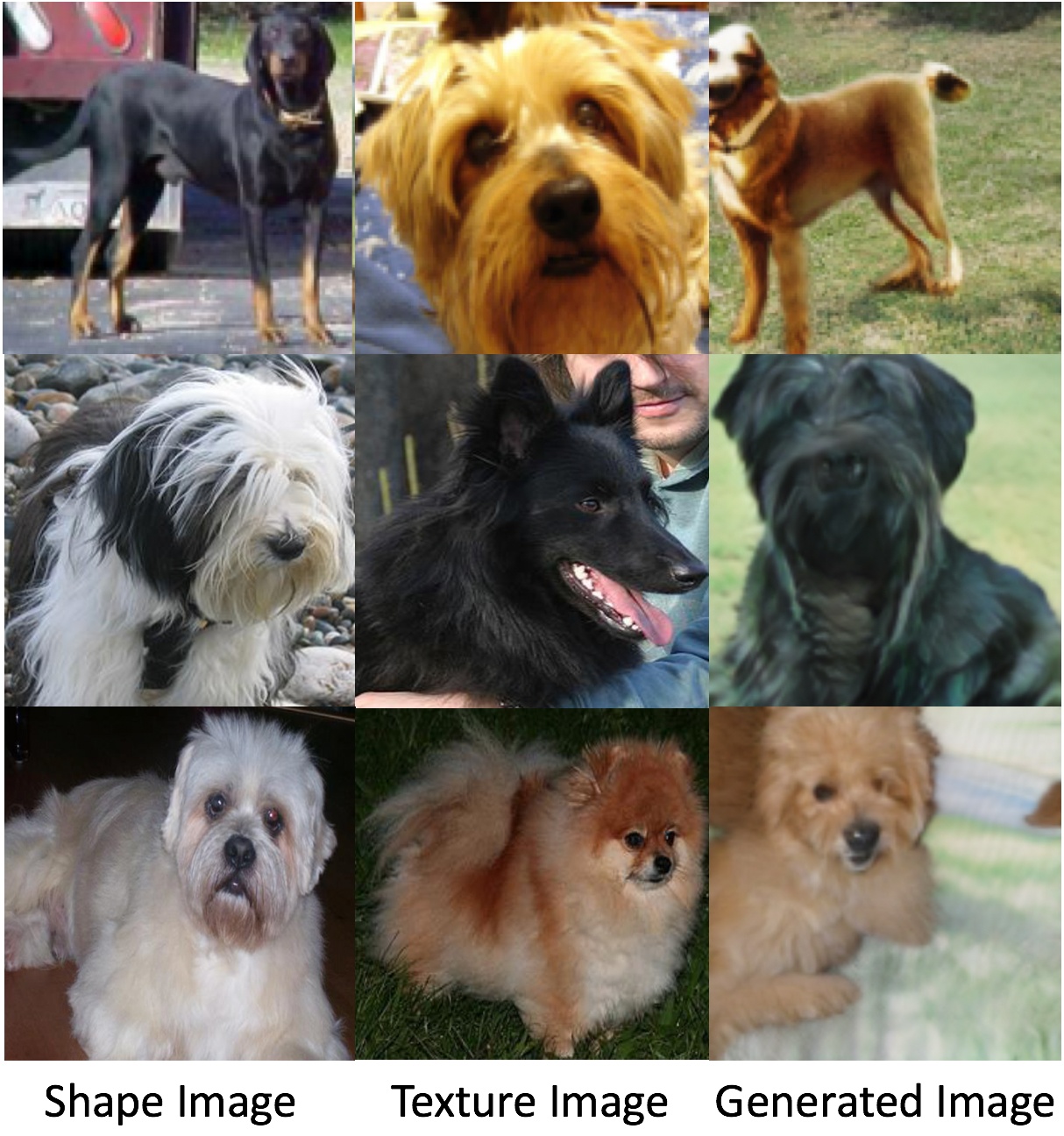}
    \captionsetup{font=large,labelfont=large}
    \caption{Additional results over the Stanford dogs dataset.}
    
    \label{fig:results_add_dogs}
\end{figure}

\begin{figure}[htp]
    \centering
    \includegraphics[width=13.5cm]{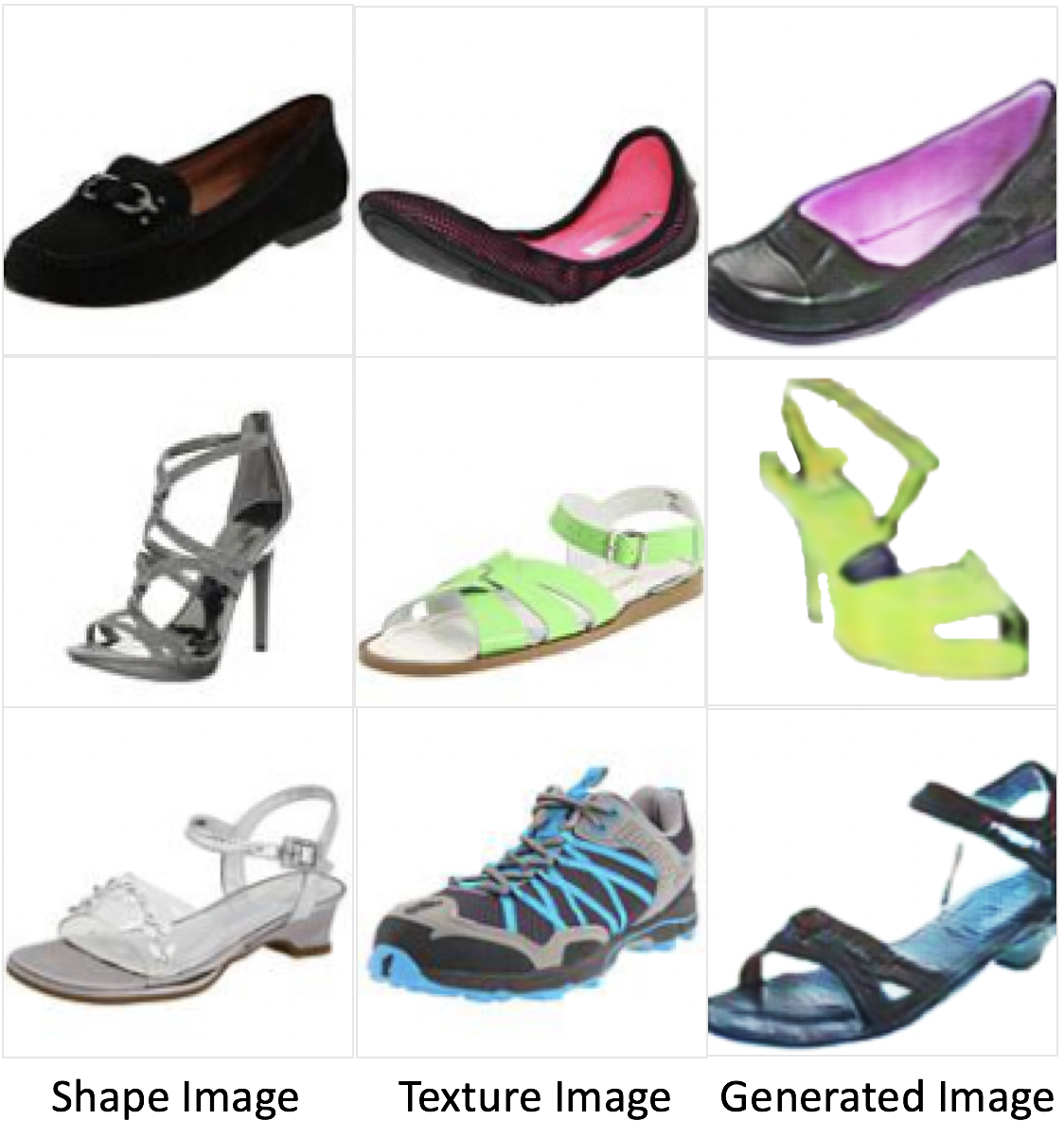}
    \captionsetup{font=large,labelfont=large}
    \caption{Additional results over the UT Zappos50k dataset.}
    
    \label{fig:results_adds_shoes}
\end{figure}

We show additional results over the three datasets in Figures \ref{fig:results_add_cub}, \ref{fig:results_add_dogs} and \ref{fig:results_adds_shoes}.

\end{document}